\documentclass[runningheads]{llncs}
\usepackage{url}
\usepackage{graphicx}
\usepackage{soul}
\usepackage{color}

\usepackage{amsmath}

\begin{document}

\title{Complexity-based partitioning of CSFI problem instances with Transformers}


\author{L. Benedetto\inst{1}, P. Fantozzi\inst{2} and L. Laura\inst{3}}
%
\authorrunning{Benedetto, Fantozzi and Laura}

\institute{
Politecnico di Milano, Milan Italy\\
\and
Dip. di Informatica e Sistemistica Universit\`a di Roma "La Sapienza", Rome Italy.\\
\and 
International Telematic University Uninettuno, Rome, Italy\\
E-mail: \texttt{luca.benedetto@polimi.it, paolo.fantozzi@uniroma1.it, luigi.laura@uninettunouniversity.net}}

\maketitle              
\begin{abstract}
In this paper, we propose a two-steps approach to partition instances of the Conjunctive Normal Form (CNF) Syntactic Formula Isomorphism problem (CSFI) into groups of different complexity.
First, we build a model, based on the Transformer architecture, that attempts to solve instances of the CSFI problem.
Then, we leverage the errors of such model and train a second Transformer-based model to partition the problem instances into groups of different complexity, thus detecting the ones that can be solved without using too expensive resources.
We evaluate the proposed approach on a pseudo-randomly generated dataset and obtain promising results.
Finally, we discuss the possibility of extending this approach to other problems based on the same type of textual representation.

\keywords{Artificial Intelligence \and Natural Language Processing \and CNF \and Computational Complexity}
\end{abstract}
\section{Introduction}
In most domains, from theoretical ones to practical ones, there is the need to cope with tasks of high computational complexity.
Even if computers can help to automatize many activities, some of these activities cannot be solved easily using an algorithm, because a subset of their instances (if not all) require computational resources that are not available or too expensive.
Usually, the solution to this issue consists of avoiding automation at all, even when most of the problem instances are easily solvable, and only few of them are computationally unfeasible. 
A better solution could be to classify the instances of the problem, partitioning them depending on the complexity of their solution, which would enable to tackle the feasible ones.

To provide this type of partitioning for a given problem, we propose a supervised machine learning approach made of two phases.
The first step consists in training a neural network model to (try to) solve the problem.
At this stage, the choice of the neural architecture to use is a critical point, because the complexity of the architecture will correspond to the boundary of the partitioning.
Indeed, since a neural network is just an algorithm with a certain computational complexity, it is not possible to solve instances that require higher computational complexity with respect
to the neural network. 
The output of the first step is made of instances of the problem that are correctly solved and instances that are not solved by the model.
We then label the instances that were not solved as ``hard'' and the other instances as ``easy''.
Finally, in the second phase, we use these new labels as target values to train another model, whose architecture is identical to the model trained in the first phase. 
After training, the second model will be able to identify -- with a certain amount of confidence -- if an instance of the original problem is either easy or hard to be solved.

To evaluate the proposed approach, we experiment on the Conjunctive Normal Form (CNF) Syntactic Formula Isomorphism problem (CSFI), trying to build a model to classify the complexity of the isomorphism checking on a pair of CNF. 
An example of a CNF is the following:
\begin{equation*}
    (a\vee b)\wedge (\neg a\vee c\vee\neg b)\wedge (\neg c\vee c)
\end{equation*}

As for the models, we use the Transformer architecture \cite{vaswani2017attention}, which is one of the most complex neural network architectures available.
Specifically, since the Transformer has a $O(n^2)$ complexity, we know that it cannot solve problems more complex than that. 
\section{Related work}
The \emph{CNF Syntactic Formula Isomorphism} (CSFI) problem is the task of stating whether two CNF Boolean formulas are semantically isomorphic \cite{AT96,AT00}. 
This problem is also related to the formula equivalence problem: indeed, if two formulas are semantically isomorphic then they are also semantically equivalent, but not the opposite.
Furthermore, Thierauf~\cite{Thie98} showed that the Graph Isomorphism problem (GI) is polynomial time reducible to Formula Isomorphism problem (FI), thus showing that FI is in the GI class. 
Ausiello et al.~\cite{ACFL20} proved that CSFI belongs to GI as well.

The relation between CSFI and GI is interesting also because GI has been studied in depth in the literature. 
Some important papers about Graph Isomorphism are, for instance: the classic work of McKay~\cite{McKay1981-yx}, the work from McKay and Piperno~\cite{McKay2014-gs} and the work from Babai~\cite{Babai2016-xf} that shows a quasi-polynomial time algorithm for GI.

In \cite{nontrivcnf}, the authors define two classes of non-isomorphic formulas (i.e. trivial and non-trivial) based on the properties of the formulas. 
This classification is useful to partition CNFs in different complexity classes, since the paper describes a technique to check the non-isomorphism of two formulas (under some conditions) that has different complexity depending on the class of the formulas.

In the literature, there are some attempts to build neural network models to solve classical theoretical computer science problems. 
In \cite{selsam2018learning} the authors try to train a Message Passing Neural Network (as defined in \cite{pmlr-v70-gilmer17a}) to solve instances of the Boolean satisfiability problem. 
Also, in \cite{prates2019learning}, the authors train a Graph Neural Network to solve instances of the Traveling Salesperson Problem.
To the best of our knowledge, no previous work proposed the usage of Transformers (nor other types of neural networks) to estimate the complexity of the instances of a given problem.

\subsection{Transformers}\label{subsec:transformers}

The Transformer is an architecture originally proposed for sequence to sequence tasks in Natural Language Processing (NLP).
An example of a sequence to sequence task is machine translation, where sentences from one language are converted to sentences in another language.
Since its introduction in 2017 \cite{vaswani2017attention}, the Transformer enabled a fast-paced improvement of the state of the art performance in several NLP tasks (not only sequence to sequence), and many powerful models are based on this architecture (e.g. BERT \cite{devlin2019bert}, XLNet \cite{yang2019xlnet}, GPT-3 \cite{NEURIPS2020_1457c0d6}).
The main technological innovation of Transformers was the usage of a self-attention mechanism, which enables the model to extract features for each word to figure out how important all the other words in the sentence are with respect to the aforementioned word, and this capability enables the Transformer to more accurately model natural language and the relationships between different words.
Currently, Transformer-based models are commonly used in NLP tasks in many different domains -- such as education \cite{benedetto2021application}, and recommender systems \cite{penha2020does} -- and successfully applied in computer vision \cite{parmar2018image,carion2020end}.

In this paper, we take inspiration from previous research and use Transformers for a task not related to NLP, proposing a model capable of partitioning CNF in complexity classes.
Specifically, we implement a modified version of RoBERTa \cite{liu2019roberta}, which is a model originally built upon BERT with some key modifications to the pretraining objective and the hyperparameters, which make it more adequate to the task we have at hand.
Crucially, RoBERTa uses BPE \cite{sennrich2016neural}, a byte-level tokenizer, to convert the input sequences into tokens while BERT uses the WordPiece tokenizer \cite{schuster2012japanese}.
Since we are not really dealing with language but CNFs (made of literals) a byte-level tokenizer is more appropriate.
As for the architecture, the original RoBERTa model is a multi-layer transformer model which is pre-trained on the task of Masked Language Modelling and can later be fine-tuned on several downstream tasks.
Masked Language Modelling consists in masking one of the input tokens with a special token and asking the model to fill the blank.
Fine-tuning on a downstream task consists in adding one or more layers on top of the network in order to adapt it to the task at hand.
In this paper, we deal with a sentence classification task, therefore we add a fully connected classification layer on top of the original network.
Also, since the task of MLM is not adequate for CNFs, we directly train the whole model on the sentence classification task, in order to leverage the attention mechanisms of the original RoBERTa model.

\section{A Neural Network approach for complexity partitioning}

As described in the introduction, the proposed approach is made of two steps. 
First, i) we train a model to classify pairs of CNF as isomorphic or non-isomorphic (separately considering trivial and non-trivial non-isomorphism), then ii) we train a separate model -- with the same architecture -- to predict the complexity a given pair of CNF, using as gold standard the correctness of the prediction of the first model.
The output of the second model indicates, with a given confidence, whether that specific instance of the original problem (i.e. CSFI for the given pair of CNF) is either easy or hard to solve.





In the CSFI problem, each instance is made of a pair of CNF, which are given as input sequence to the model.
Since our models cannot directly work on the CNF, the input sequence must be tokenized.
Tokenizing a sequence, in our case a CNF, consists of splitting it into smaller components, which are then converted to ids through a look-up table.
As an example, a way of tokenizing a natural language sentence is to split it in words and associate an id to each word.
As we are not dealing with words but rather with CNF, we use a byte-level tokenizer: specifically, we use the Byte Level BPE Tokenizer, in its implementation from \textit{HuggingFace} \footnote{\url{huggingface.co/docs/tokenizers/python}}.
In practice, the tokenizer separates all the literals and the operators, as in the following example: $a \vee c \wedge c\vee b$ is tokenized into [$<$s$>$, $a$ $\vee$, $c$, $\wedge$, $c$, $\vee$, $b$, $<$/s$>$], where ``$<$s$>$'' and ``$<$/s$>$'' are two special tokens that indicate the beginning and the end of the CNF, and then converted into a list of IDs.
We use 100 as vocabulary size (\texttt{vocab\_size=100}), which is sufficient for the CNFs considered in this study, and \texttt{min\_frequency=1}, meaning that all the literals are considered (even if they appear only once) and no ``unknown'' token is used.
The same tokenizer is used in both phases of the proposed approach.

\subsection{Model}

For the two phases of our approach -- i) CSFI and ii) complexity partitioning of instances of CSFI -- we use two different models.
However, these two models have the same architecture, since in both phases the task is sequence classification; the only difference (but a crucial one) is the inner weights of the neural networks, which are learned at training time.

Both models are based on RoBERTa \cite{liu2019roberta}, and we do not perform any modification to its.
That is, it is made of 6 hidden layers, we use 12 attention heads, and we set 512 as maximum input length\footnote{we use the \texttt{RobertaConfig} and \texttt{RobertaForSequenceClassification} classes from the HuggingFace transformers library (\url{https://huggingface.co/transformers})}.

However, we add a dense layer and a linear output layer on top of that, in order to adapt RoBERTa to the task of sequence classification.
Also, since each model receives as input pairs of CNF (namely $\alpha$ and $\beta$) and we do not want the ordering of the pair to influence the outcome of the classification, we consider in parallel the two possible concatenations ($\alpha\beta$ and $\beta\alpha$).
In practice, we retrieve the output of the RoBERTa model for the two sequences, concatenate it into a unique array, and give it as input to the classification layers (i.e. dense layer and linear output layer).

\section{Experimental results}
In this section we present and analyse the experimental results obtained to evaluate the proposed architecture. 
The hardware configuration used for the experiments is the following: Intel i7-9700k with 8 cores with frequency of 3.60 Ghz, 32 GB of DDR4 RAM, a NVIDIA Titan Xp graphic card with 12 GB of RAM and CUDA libraries version 11.3.

\subsection{Dataset}
All the datasets used for the experiments have been generated using the algorithms defined in \cite{nontrivcnf}, implemented with the python library presented in the same work. 
All the pairs of isomorphic CNFs are composed by i) a randomly generated formula and ii) a formula built by applying a random isomorphism to the first one. 
All the pairs of non-isomorphic formulas are composed
by i) a randomly generated CNF $\alpha$ and ii) a formula $\gamma$ obtained by applying a first step to transform $\alpha$ to another CNF $\beta$ that is not isomorphic to $\alpha$, and then a second step to build $\gamma$ as isomorphic with respect to $\beta$.
Since, by definition, the syntactic isomorphism property is transitive, in each pair defined as before, $\gamma$ is not isomorphic to $\alpha$.

The random generation of all the CNFs in the datasets respects the following parameters:
\begin{itemize}
    \item the number of symbols in each formula is in the range $[15,25]$;
    \item all the symbols in the datasets are randomly chosen from same set of 25 symbols;
    \item the number of clauses in each formula is in the range $[10,15]$;
    \item all the clauses have the same cardinality (8);
    \item all the formulas are monotone (there are no negated literals).
\end{itemize}

Given a CNF, the trivial non-isomorphic generation is performed by using one random modification between the following:
\begin{itemize}
    \item adding a new clause with the same cardinality randomly choosing symbols from the ones in the CNF;
    \item adding a symbol (from the ones in the CNF) to an existing clause, thus changing its cardinality;
    \item adding a symbol that is not present in the CNF.
\end{itemize}

The dataset used to train the first model is composed of $50^\cdot 000$ pairs of CNFs, belonging to three categories: 50\% pairs are syntactically isomorphic, 25\% are trivially non-isomorphic, and 25\% are non-trivially non-isomorphic.
The pairs have been labeled with 0 or 1 for non-isomorphic and isomorphic pairs, respectively.
The evaluation and test datasets have been generated using the same parameters of the train dataset, with the only difference that they are made of $10^\cdot 000$ pairs each.

The train dataset for the second model is built with the following steps. 
First, we build a dataset which has the same characteristics of the datasets used to train and evaluate the first model.
Then, we use the first model to classify the samples of this dataset, and compare the true label with the predicted class, creating a new label that indicates whether the first model correctly classified each pair of CNFs (0 indicating correct classification and 1 wrong classification).
After that, we randomly select a subset of the samples such that 50\% of the samples are correctly classified and 50\% are wrongly classified. 
The actual size of this dataset will depend on the size of the smallest subset between the classified samples.

Both the evaluation and test datasets for the second model are generated following the same guidelines of the previous datasets to obtain $10^\cdot 000$ pairs for each dataset, 50\% of the trivially non-isomorphic and 50\% non-trivially non-isomorphic.
Since we use these datasets to test the second model, which has been trained to classify the samples that are difficult to solve, we can use only non-isomorphic pairs, because we have no information about the difficulties of the isomorphic pairs.

\subsection{First phase results}

\begin{table}[ht]
    \centering
    \begin{tabular}{c|c|c|c}
         Accuracy & Precision & Recall & F1 \\ \hline
         0.730 & 0.669 & 1.0 & 0.801
    \end{tabular}
    \vspace{0.5cm}
    \caption{Evaluation metrics for the first model on the test dataset.}
    \label{tab:1st-results}
\end{table}

Table \ref{tab:1st-results} shows the results obtained on the test set with the first model; we use accuracy, precision, recall, and f1-score as evaluation metrics.
The training lasted 10 hours and 30 minutes.

\begin{figure}[ht]
    \centering
    \includegraphics[width=0.8\textwidth]{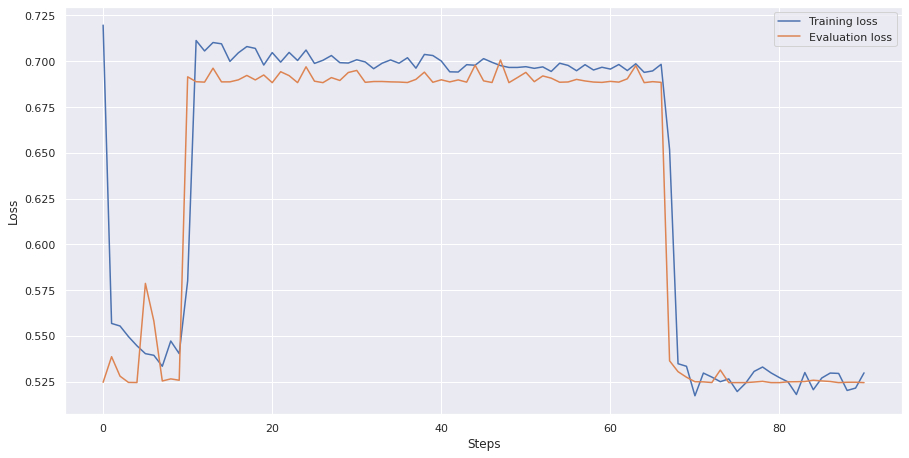}
    \caption{Training loss and evaluation loss for the training of the model for the 1st phase.}
    \label{fig:1st-model-loss}
\end{figure}

\begin{figure}[ht]
    \centering
    \includegraphics[width=0.8\textwidth]{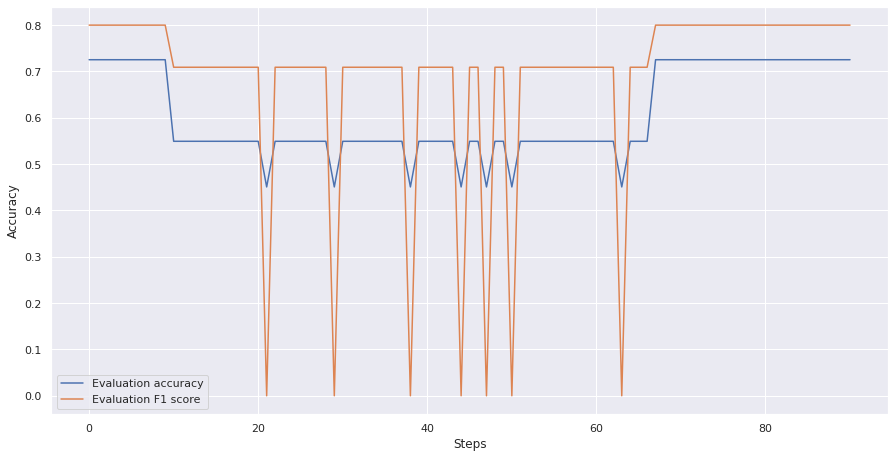}
    \caption{Evaluation accuracy and F1 score for the training of the model for the 1st phase.}
    \label{fig:1st-model-accuracy-f1}
\end{figure}

As shown in the loss plot in the Figure \ref{fig:1st-model-loss}, the model starts to learn very well in the first
epoch. 
Then in the second epoch something changes in its ability to generalise, as if it was starting to learn from scratch again. Finally, at the end of the sixth epoch, there is a major learning step, and the loss stabilises, reaching lower loss values than at the beginning.

Both the evaluation accuracy and the evaluation F1-score follow the same trend as well, as
shown in Figure \ref{fig:1st-model-accuracy-f1}. Indeed, the two curves have the same discontinuities at the end of the second and the sixth epoch, reaching, after that, the best results.

\begin{table}[ht]
    \centering
    \begin{tabular}{l|c|c|c}
         & iso. & trivial non-iso. & non-trivially non-iso. \\ \hline
         right label& 4986 & 1704 & 0 \\
         wrong label& 0 & 0 & 2462 \\
    \end{tabular}
    \vspace{0.5cm}
    \caption{Number of samples that are correctly labeled by the first phase model.}
    \label{tab:1st-wrong}
\end{table}

After training the first model, the training dataset for the second model was created, performing the classification of each pair of CNFs and comparing it with the ground truth values.
From the results shown in Table \ref{tab:1st-wrong}, we can notice that the only misclassified samples are the samples generated as non-trivial non-isomorphic, which are the most complex ones.

\subsection{Second phase results}

\begin{table}[]
    \centering
    \begin{tabular}{l|c|c|c|c}
         Threshold & Accuracy & Precision & Recall & F1 \\ \hline
         0.50 & 0.993 & 0.989 & 1.0 & 0.994 \\
         0.66 & 0.976 & 0.962 & 1.0 & 0.980 \\
         0.51 & 0.993 & 0.989 & 1.0 & 0.994 
    \end{tabular}
    \vspace{0.5cm}
    \caption{Results of the model for the 2nd phase applied to the test dataset. The column threshold contains
             the probability threshold of consider a sample ``easy to solve''. The first value is just the
             majority, the second is chosen to indicate a threshold that is largely bigger than the majority,
             while the last one is the best threshold that could be picked for this dataset (used only to
             compare the results).}
    \label{tab:2nd-results}
\end{table}

In the second phase we assign to each sample of the second training dataset 1 if they are misclassified
from the first model and 0 otherwise. This model should be able to recognize if a sample is ``easy to solve'',
so, after the training, we used a test dataset composed by 50\% of ``easy'' samples and 50\% of ''hard`` samples.
The results from the linear output layer of the model are transformed using a softmax to have a probability
distribution over the two labels that sum to 1. The results, after a training of 5 hours and 38 minutes,
with respect to the test dataset are shown in table \ref{tab:2nd-results}. 

We tried to apply different thresholds to the probabilities returned from the model. Indeed, in our idea,
the model should be very confident before classifying a sample as ``easy''. So we don't want to use just
the higher probability between 0 and 1, and we think that the probability to be an easy sample should be
at least the double with respect to the opposite classification. Since that the probabilities sum to 1,
then it is equivalent to say that the probability of the ``easy'' classification should be greater than
0.66. In the table of results we show: the results for just the majority in probabilities, the results
for a threshold of 0.66 and also the best results obtainable moving the threshold.

\begin{figure}[ht]
    \centering
    \includegraphics[width=0.8\textwidth]{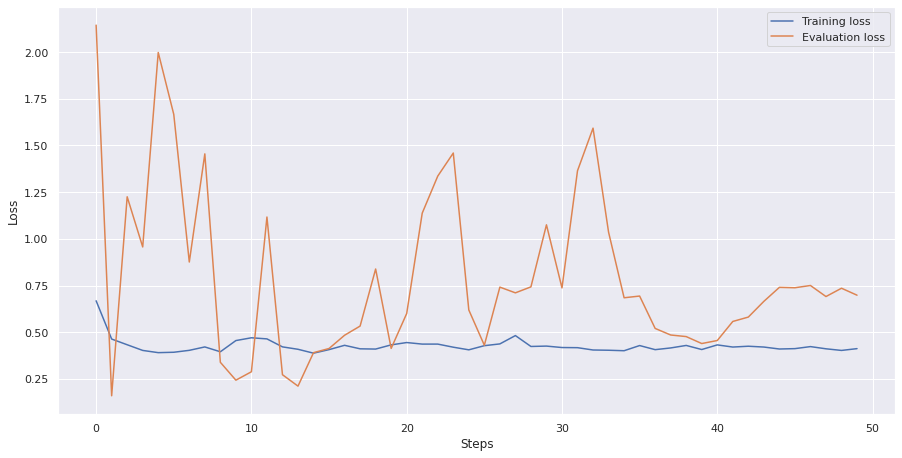}
    \caption{Training loss and evaluation loss for the training of the model for the 2nd phase}
    \label{fig:2nd-model-loss}
\end{figure}

In the loss curve plot shown in figure \ref{fig:2nd-model-loss} we can see that the model learns quickly
since that from the second epoch the training loss remains more or less stable. The evaluation loss, instead,
shows much more variance, but it becomes more stable during the epochs to become, eventually, comparable with
the training loss.

\begin{figure}[ht]
    \centering
    \includegraphics[width=0.8\textwidth]{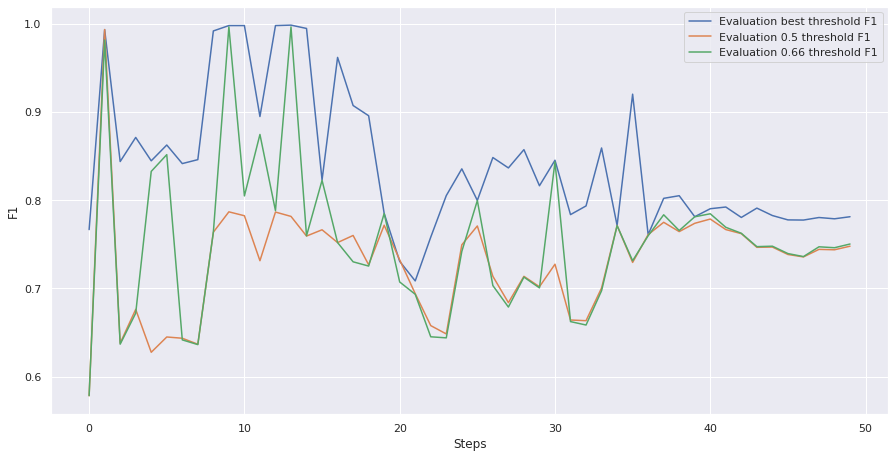}
    \caption{Evaluation F1 score for the training of the model for the 2nd phase}
    \label{fig:2nd-model-f1}
\end{figure}

The same evaluation performed on the results on the test dataset, is also performed continuously on the
evaluation dataset during the training. In the figure \ref{fig:2nd-model-f1} can be seen the F1 score
for the 3 different thresholds described before. It is easy to notice that the gap between them tends
to be more stable with the epochs. At the end, the gap seems to be small, and this is also confirmed
in the results shown before.

\begin{figure}[ht]
    \centering
    \includegraphics[width=0.8\textwidth]{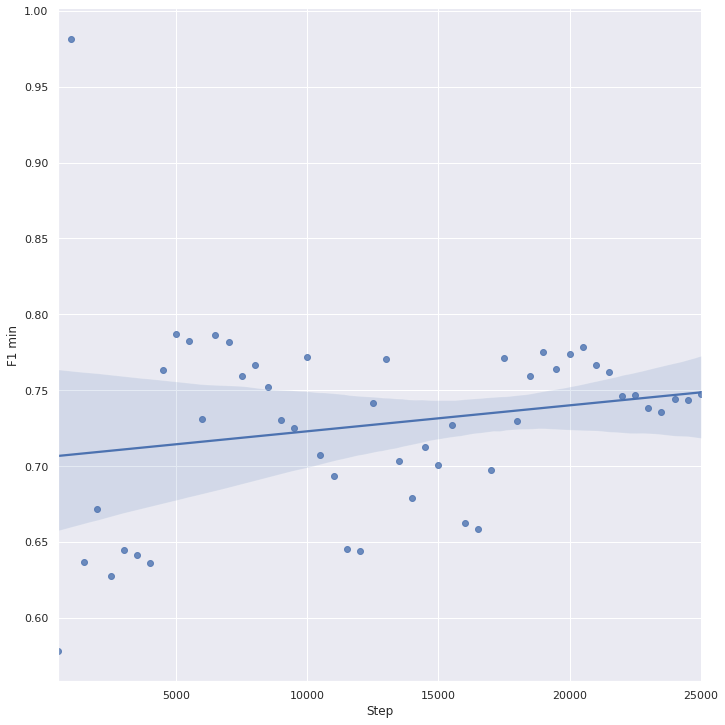}
    \caption{Minimum evaluation F1 score, in the range [0.5,1.0] for the training of the model for the 2nd phase}
    \label{fig:2nd-model-min-f1}
\end{figure}

\begin{figure}[ht]
    \centering
    \includegraphics[width=0.8\textwidth]{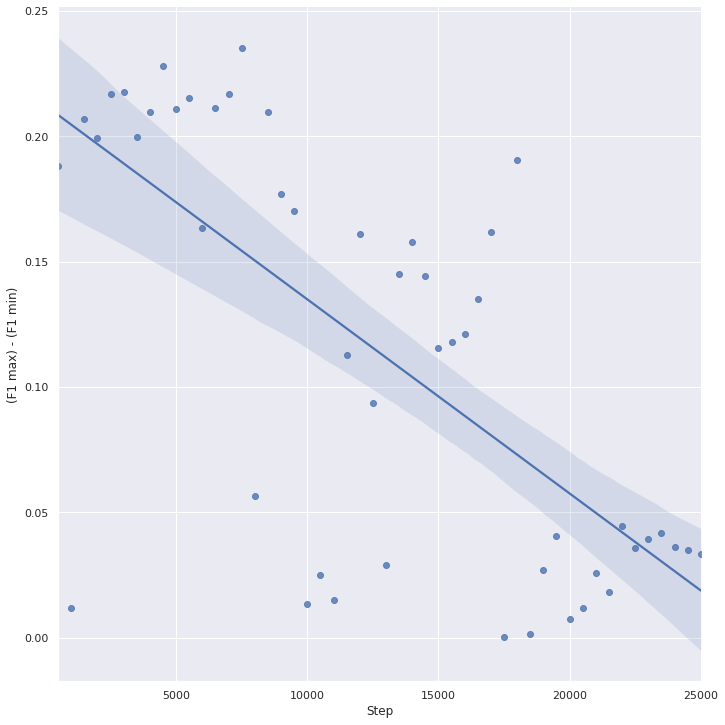}
    \caption{Plot of the difference between the maximum F1 score (obtained with the best
    threshold) and the minimum F1 score (obtained with a different threshold) during
    the training of the 2nd model.}
    \label{fig:2nd-model-range-f1}
\end{figure}

We also show, in the figure \ref{fig:2nd-model-min-f1}, the curve of the F1 score that is the mininum
between the scores obtained by using the 3 different thresholds during the training. We can see that the
trend of the minimum F1 score is a slowly increasing. At the same time, the gap between the minimum
and the maximum F1 score given by the different thresholds, decrease quickly during the epochs as shown
in figure \ref{fig:2nd-model-range-f1}.

\begin{figure}[ht]
    \centering
    \includegraphics[width=0.8\textwidth]{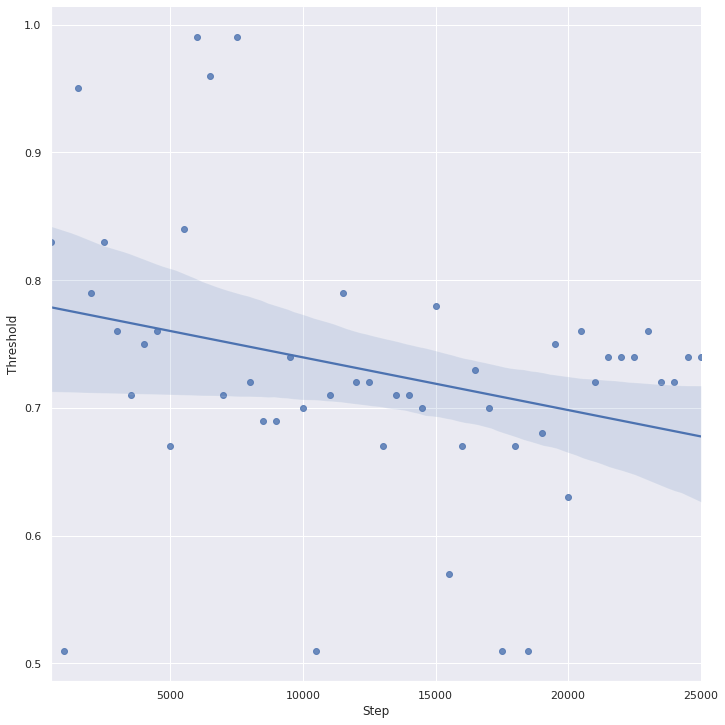}
    \caption{Plot of the threshold on probabilities that gives the best results on the evaluation dataset
             during the training.}
    \label{fig:2nd-model-thresholds}
\end{figure}

It is also interesting to analyse the trend of the thresholds computed during the training. They are
computed looking for the threshold on the probabilities that returns the best results overt the
evaluation dataset. It can be seen in the figure \ref{fig:2nd-model-thresholds} that the threshold
tends to decrease, and so the model becomes more accurate to estimate the probabilities.

\section{Conclusion}
In this paper, we proposed a neural network-based approach to partition instances of the CNF Syntactic Formula Isomorphism problem (CSFI) into different complexity classes.
From this categorization, it is also possible to get a complexity threshold, which separates the easy instances (under the threshold) and the instances for which it is hard to find a solution (above the threshold).

The proposed approach is made of two neural models -- both based on the Transformer architecture -- which serve different purposes.
The first one is trained to attempt to solve instances of the problem; however, rather than using its predictions as solutions to the instances, we leverage them to build a training dataset.
Specifically, we build a dataset made of two classes (easy and hard), and use it to train the second model, which then learns to partition CNFs in the two complexity groups.

The advantage of this approach is that it can be used to identify the instances of a problem whose solution could be automatized without needing too expensive hardware resources, allowing the planning of different solution strategies depending on the cost of the solution itself.

Finally, it is worth noting that in this paper we only experimented on CNFs and instances of the CSFI problem, but this approach could most likely be extended to different problems.
Indeed, we only leverage the textual representation of the CNFs and other problems (such as the Boolean satisfiability problem) have similar representations.
Future research will work towards that direction.

\noindent \textbf{Acknowledgements} 
Luigi Laura gratefully acknowledges the support of NVIDIA Corporation with the donation of the Titan Xp used for this research.

\newpage
%
%
%
\bibliographystyle{splncs04}
\bibliography{bibliography}
\end{document}